\documentclass[11pt]{article}
\usepackage[utf8]{inputenc}
\usepackage{verbatim}
\usepackage{url}
\usepackage{amsmath}
\usepackage{graphicx}
\usepackage[a4paper]{geometry}
\usepackage{setspace}
\usepackage[authoryear]{natbib}
\makeatletter
\newcommand{\lyxaddress}[1]{
	\par {\raggedright #1
	\vspace{1.4em}
	\noindent\par}
}

\usepackage{authblk}

\graphicspath{figures/}

\usepackage[left,modulo]{lineno}

\makeatother

\begin{document}
\title{Mistake gating leads to energy and memory efficient continual learning}
\author{Aaron Pache$^{2}$ and Mark CW van Rossum$^{1,2}$}
\maketitle

\lyxaddress{1 School of Psychology\\
 2 School of Mathematical Sciences\\
 University of Nottingham, Nottingham NG7 2RD, United Kingdom}

\begin{abstract}
Synaptic plasticity is metabolically expensive, yet animals continuously
update their internal models without exhausting energy reserves. However,
when artificial neural networks are trained, the network parameters
are typically updated on every sample that is presented, even if the
sample was classified correctly. Inspired by the human negativity
bias and error-related negativity, we propose `memorized mistake-gated
learning'---a biologically plausible plasticity rule where synaptic
updates are strictly gated by current and past classification errors.
This reduces the number of updates the network needs to make by $50\%$
to $80\%$. Mistake gating is particularly well suited 1) for  incremental
learning where new knowledge is acquired on a background of pre-existing
knowledge, and 2) for online learning scenarios when data needs to
be stored for later replay. Here,  mistake-gating strongly reduces
the required storage buffer size. Moreover,  mistake-gating increases
robustness when learning from imbalanced classes. The algorithm can
be implemented in a few lines of code, adds no hyper-parameters, and
comes at negligible computational overhead. Learning on mistakes is
an energy efficient and biologically relevant modification to commonly
used learning rules that is well suited for continual learning.
\end{abstract}
Experimental evidence suggests that biological learning is a metabolically
costly process. While the reasons are currently not fully understood,
these costs can be substantial. For instance, flies subjected to aversive
conditioning and subsequently starved, died 20\% sooner than control
flies \citep{Mery2005b}. Similarly, flies doubled sucrose consumption
after long term memory formation \citep{Placais13}, and memory can
be boosted by mitochondrial stimulation \citep{amrapali2026mitochondrial}.
This points to an evolutionary imperative to learn frugally, that
is, with minimal updates to the brain's networks \citep{pache2023energetically}.

Indeed, humans don't learn equally from every sample, but place far
more importance on mistakes and prediction errors, a phenomenon known
as the negativity bias \citep{Rozin2001}. A familiar example is the
`jolt' we experience when typing a word incorrectly. When such mistakes
occur, a large event-related potential known as the error-related
negativity (ERN) is elicited in the EEG, which usually precedes behavioural
changes like post-error slowing and post-error improvement in accuracy
\citep{Kalfaoglu2017}. Indeed, subjects learn better on stimuli that
evoke a larger EEG \citep{de2020processing}. Moreover, the ERN has
been linked to dopamine signalling \citep{Holroyd2002} - a plasticity
modulator associated with persistent forms of synaptic plasticity
(late-phase LTP) \citep{OCarroll2004}.

In contrast, most training algorithms for artificial neural networks
update their parameters on all samples, even if classified correctly.
In particular for online continual learning, such as natural vision
where the sample variation is virtually limitless, this appears excessive.
First, late in learning most samples will be correctly classified
and might be skipped; second, if concrete samples need to be stored
for later processing, focusing on the mistakes reduces offline storage
buffer requirements.

Previous work has found that for many datasets, it suffices to learn
from a small subset of critical samples termed the `core-set' \citep{Agarwal2005}.
However, finding the core-set is hard  \citep{toneva2018empirical,8576015,DBLP:journals/corr/abs-1910-00762,paul2021deepdiet,coleman2020selectionproxy}.
More advanced algorithms seek to filter out incorrect and task irrelevant
samples \citep{mindermann2022prioritized}. Some might require another
model trained beforehand to identify the coreset. For a new, unseen
dataset, it won't always be clear what the core-set is. As a `softer' alternative, importance sampling relies more on certain
samples than others during training \citep{pmlr-v80-katharopoulos18a}.

One ideally would have a simple algorithm that automatically finds
important samples in an online setting, which is particularly relevant
for biological and continual learning. To better understand biological
learning as well as develop more human-like AI, we explore whether
mistake gating can reduce putative metabolic cost of learning and
memory buffer requirements.

\section*{Results}

\subsection*{Mistake-gated learning}

\begin{figure}
\centering{}\includegraphics[width=18cm]{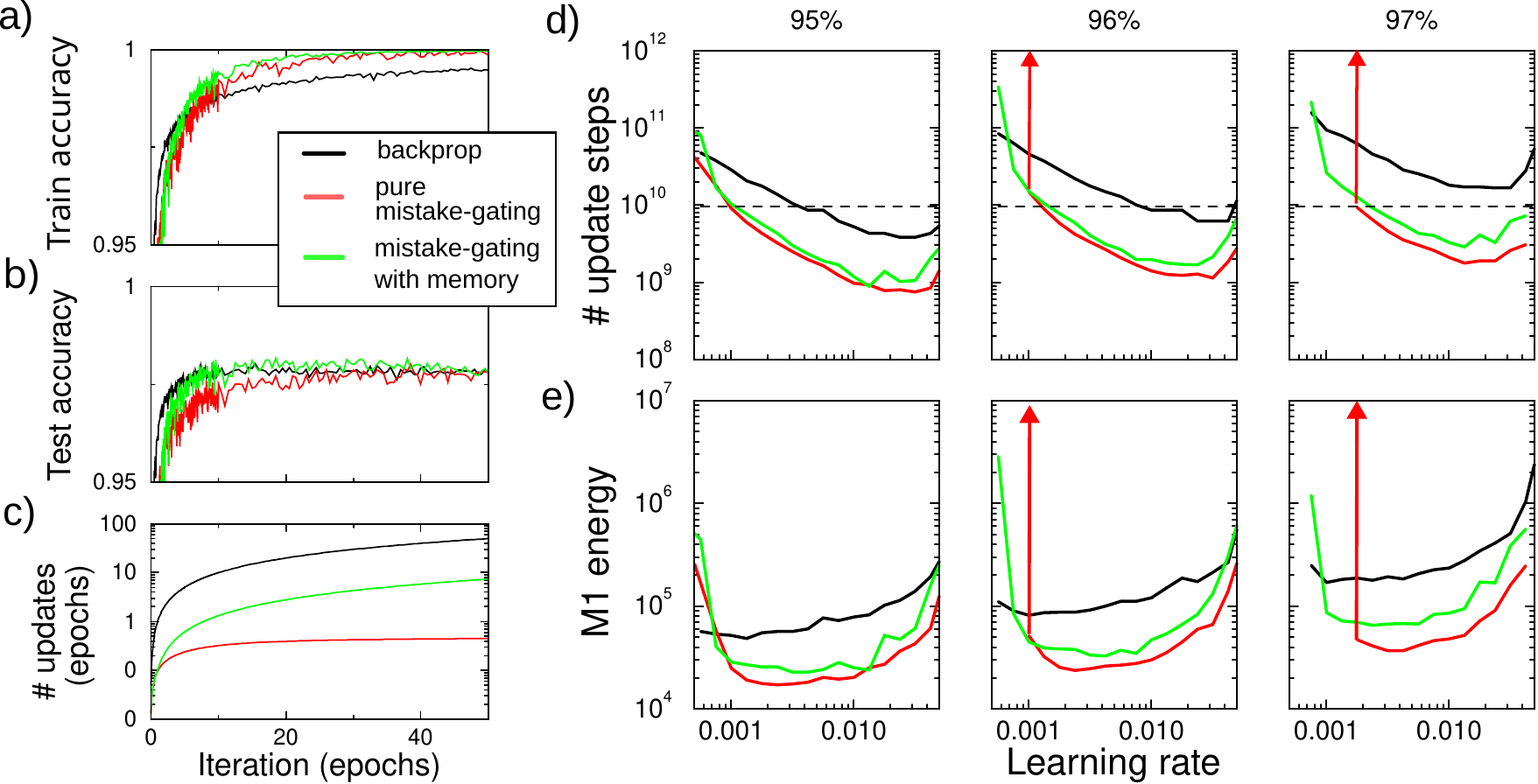}\caption{Comparison of standard backpropagation  to mistake-gated learning
rules on the MNIST digit classification task. Standard backpropagation
(black), pure mistake gated learning which only update when a mistake
is made (red), and mistake-gated learning which updates when the sample
is misclassified or was misclassified in the past (green; the focus
of this study). \protect \\
a) Training accuracy as a function of the number of training cycles.
Accuracy improves more rapidly with mistake gated learning rules compared
to backprop, despite using fewer updates. \protect \\
b) Test-set accuracy grows more slowly than training accuracy for
mistake gating. \protect \\
c) The total number of parameter updates (normalized by network size
and dataset size). Backprop requires many more parameter updates than
either mistake gating rule.\protect \\
d) The number of backpropagation steps required to reach certain test
accuracy (95\%, 96\% and 97\%) as a function of learning rate. An
intermediate learning rate is optimal for all algorithms. Both pure
mistake gated (red) and with memory (green) require far fewer training
steps than backprop (black). Pure mistake-gated learning (red) can
however never reach the higher accuracies as training accuracy reaches
perfection;  with memory it can. (dashed line: number of updates over
one epoch under backprop for reference).\protect \\
e) The putative $M_{1}$ energy measure shows a similar behaviour
as the number of updates, albeit at a lower optimal learning rate.
}\label{fig:trace}
\end{figure}
We examine a supervised learning scenario. In such scenarios the network
is trained by minimizing a surrogate loss function, commonly a mean-square
error or cross-entropy loss. The reason for using a surrogate loss
is that the true loss -- whether the classification is correct or
not -- is not differentiable. While this is common practice, any
sample's surrogate loss is almost always non-zero even if classified
correctly, and weights will almost always be updated. The sample selection
methods listed above typically rely on the surrogate loss. What happens
if we instead make the update conditional on incorrect answers?

To test this we first train a feed-forward neural network to classify
digits from the well-known MNIST dataset. Throughout we use a standard
(non-convolutional) network with a single hidden layer with 200 hidden
units. After learning, when presented with a test image, the output
neuron with the highest activation should correspond to the class
label of the presented digit. We trained the network with standard
Stochastic Gradient Descent with a default learning rate of 0.01 (which
close to optimal, see Fig.\ref{fig:trace}). Typically we train to
a certain criterion accuracy on the test set. For biological realism
samples are presented one at a time, i.e. without batching (see Discussion).

A first idea is to only update the weights when a sample is misclassified.
If on the other hand the sample is correctly classified, no update
occurs and the algorithm skips to the next sample, as in the classic
perceptron algorithm \citep{Rosenblatt1958,Rimer2001,Rimer2006}.
We call this pure mistake-gating. With pure mistake-gating the training
accuracy grows rapidly compared to traditional backprop and quickly
reaches perfection, Fig.~\ref{fig:trace}a (cf.~red to black curves).
The generalization performance, as evaluated by measuring performance
on the test set, grows more slowly than traditional backprop, Fig.~\ref{fig:trace}b.
Yet, the number of update steps is still substantially less than traditional
backprop (black curve), Fig.\ref{fig:trace}c.

While pure mistake gated learning achieves a low error rate on the
training set, the generalization on the test data is fragile. To show
this we attempt to train to a certain criterion test-set accuracy.
Pure mistake gating cannot reach the highest test-set accuracies when
the learning rate is small (red curves in right-most panels of Fig.\ref{fig:trace}d).
This limits the generality and practical use of pure mistake gating.
The reason is that gating results in decision boundaries close to
the training samples, and hence a slight change in input can easily
switch the output. Instead, traditional backprop steers the network
away from the decision boundaries by updating even if the output is
correct, thereby improving generalization performance on the test-set.
When learning rates are large, mistake gating overshoots the decision
boundary, leading to more robust solution. One way to increase robustness
is to add a margin to the learning, leading to the well-known hinge
loss used in support vector machines \citep{dougan2016unified}. Biologically,
this requires knowledge of the activation of the neurons with respect
to a particular margin and metabolically, requires knowing the margin
that minimises energetic cost. Thus we concentrate on an automatic,
margin-less alternative here.

\subsection*{Memorized mistake gating}

As pure mistake-gated learning is fragile, we modified the algorithm
to memorize all incorrect samples. It updates the synapse when the
sample is currently misclassified but also when it re-encounters a
sample that was at \textit{any previous time} misclassified. We call
this \emph{memorized mistake gating. }When training with this learning
rule, it uses more update steps than pure mistake-gated learning,
Fig.~\ref{fig:trace}c (green). Yet, it is no longer fragile and
still uses only a fraction of the update steps that regular backprop
uses across the optimal range of learning rates and target criteria,
Fig.~\ref{fig:trace}d (green). Only in the limit of very small learning
rates, where all samples produce effectively a random output on the
first pass and so most samples are labeled as mistakes, memorized
mistake gating uses as many updates as backprop. %

We also measured the cumulative $L_{1}$ norm of the updates, or $M_{1}$
energy, that was proposed as an approximation for metabolic cost of
biological learning, for instance as an anabolic/catabolic measure
for the number of receptors added to and removed from the synapses
\citep{li2020energy}. Mistake gating also reduces the $M_{1}$ energy.
This is despite that the average updates in mistake gating are on
average larger than for backprop, as the incorrect samples lead to
larger gradients.

Thus while memorized mistake-gating requires a similar or larger number
of forward passes through the network, it drastically reduces the
number of the synaptic updates required to achieve a certain performance.
As this holds across learning rates (Fig.\ref{fig:trace}c+d), the
results cannot be attributed to a change in effective learning rate.
For brevity, from here on we focus on the memorized variant of mistake
gating, and loosely refer to standard, ungated training as ``backprop''.
 %

Implementation of memorized mistake gating is straightforward. It
requires a Boolean storage array with as many elements as there are
samples in the dataset. All entries are initialized to false, but
if at any time during learning a sample is misclassified, the array
element for that sample is set to true and updated on immediately;
it is never reset and every time that sample is encountered again,
the network is updated. The mistake-gating learning rule introduces
no extra parameters, and requires no tuning, the Boolean array represents
a negligible storage overhead, and is easily added to any existing
code. 

In the Appendix we compare mistake-gating to the sample selection
proposed by \citet{DBLP:journals/corr/abs-1910-00762} and find similar
performance.

\subsection*{Identity of the mistaken samples}

\begin{figure}
\centering{}\includegraphics[width=12cm]{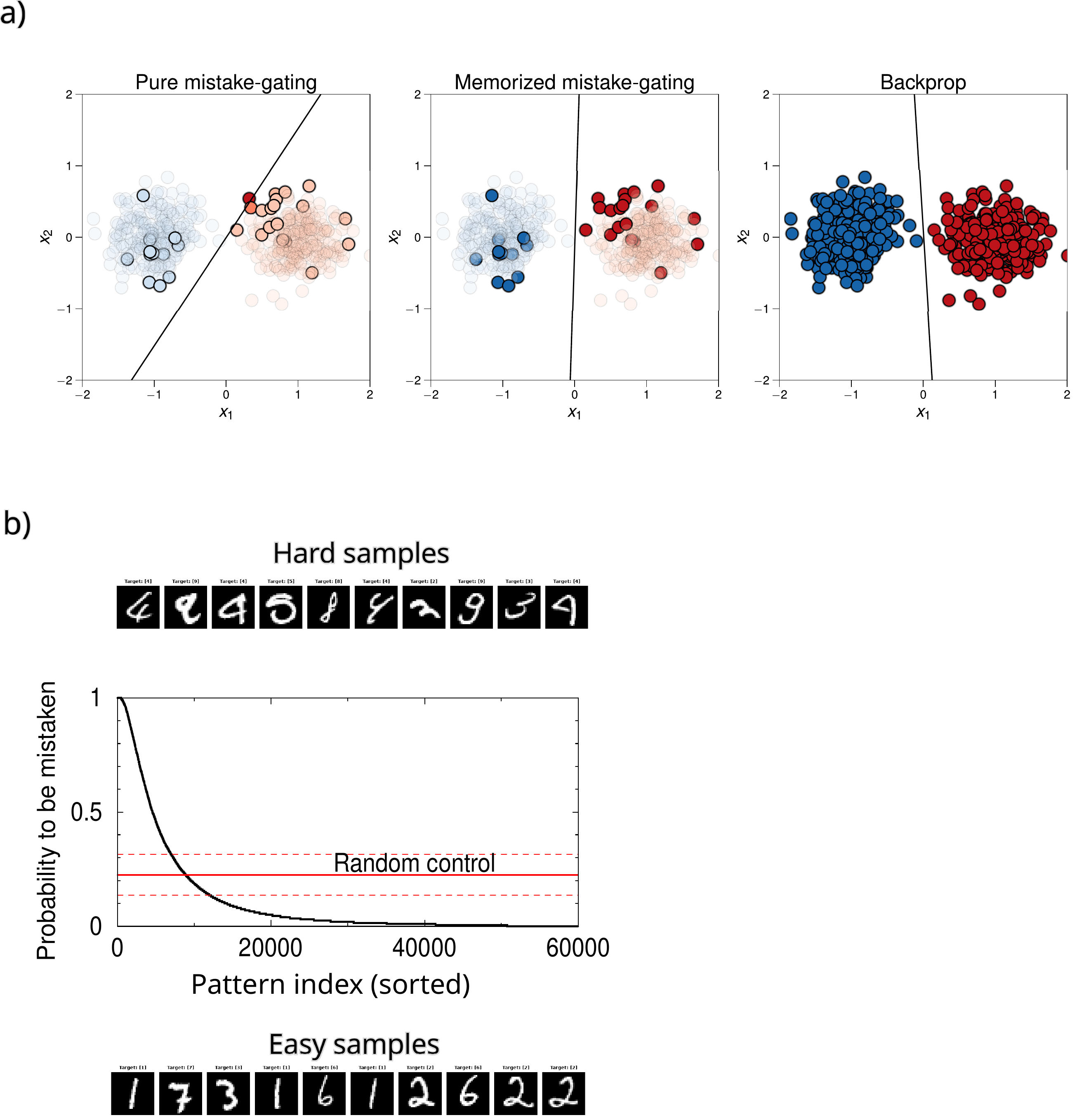}\caption{Identity of the mistake samples. a) Visualization of mistake-gating
on a simple 2D problem. Samples are colored according to how frequently
they triggered an update and unused samples are transparent. Networks
were trained for 50 epochs, with 500 samples using a small learning
rate of 0.05. Pure-mistake gating concentrates updates on samples
nearest the decision boundary; once a decision boundary is found,
learning ceases. Memorized mistake-gating distributes updates over
samples that were previously incorrect, resulting in a more robust
decision boundary. Backprop uses all samples, yielding a boundary
similar to mistake-gating but without the sparse updating.\protect \\
b) Probability to be mistaken averaged across runs. Certain samples
are harder and more likely to be mistaken than others (black; index
sorted for clarity). Uniform distribution (red) when samples would
all share the same mistake probability (dashed lines are 2$\sigma$
errors). Top samples are always mistaken, bottom samples are almost
always correct. (200 runs of MNIST, 200 hidden units, trained to 97\%
accuracy).}\label{fig:identity}
\end{figure}

We visualize each algorithm using a two-dimensional task in Fig.~\ref{fig:identity}.
Memorized mistake-gating yields a similar update sparsity as pure
mistake-gating but since it learns on previously incorrect samples
still maintains some of  the robust decision boundary of backprop.
We wondered if clustering of learning near the decision boundary translates
to high-dimensional datasets and how much a pattern's identity matters.
Generally, whether a sample is misclassified should depend on a number
of factors: 1. The network initialization. 2. When it is presented
(classification of the very first patterns will almost certainly be
wrong). 3. Whether the pattern is an outlier. We ran 200 MNIST simulations
with different initial conditions and different random pattern sequences
and tracked which patterns were memorized as mistakes, Fig.~\ref{fig:identity}b.
If the pattern identity would not matter, one would expect a flat
probability of error (red line), but the histogram is clearly skewed.
Some patterns are much more likely to be mistaken than others. The
samples that are always mistaken (top) indeed appear to be hard, while
the very rarely mistaken samples (bottom) appear easy.

\subsection*{Dynamics of mistakes}

\begin{figure}
\centering{}\includegraphics[width=14cm]{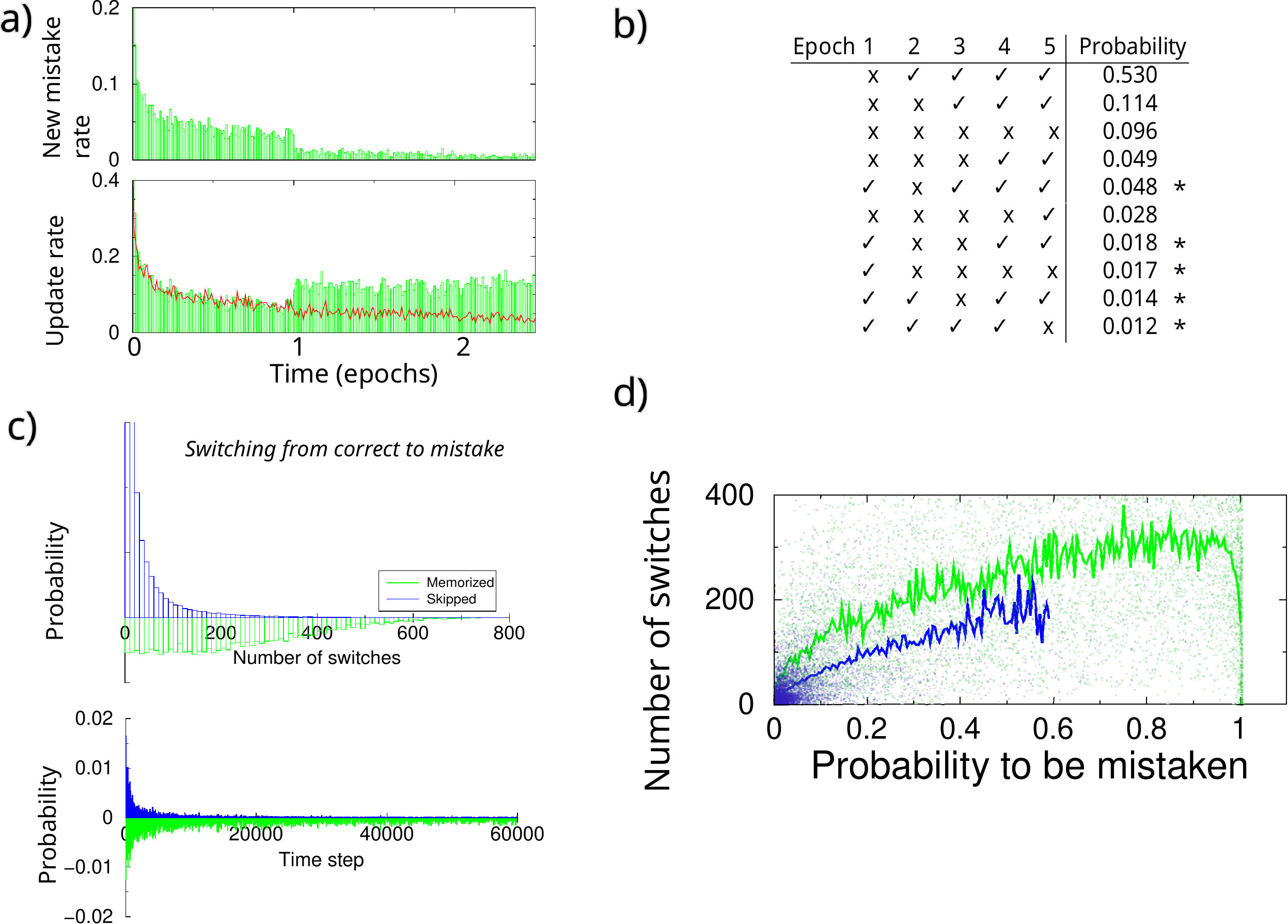}\caption{The fate of mistaken samples. a) Rate of new mistakes being memorized
(probability per time-step) and rate that the network updates. Also
shown pure mistake-gating (red).\protect \\
b) Sparsely tracking a sample's performance (that is, when it is selected
for updating). Ten most common pattern of samples being mistaken when
training on MNIST over 5 epochs. Only rarely do correct samples turn
incorrect again (stars).\protect \\
c) Exhaustively tracking sample performance at every time-step. Top:
Distribution of the number of forgetting events (switching from correct
to wrong classification). The number of switches is higher for mistakes
(green; y-axis flipped for clarity) than for skipped samples (blue).
Bottom: mistaken samples switch persist switching longer.\protect \\
d) The number of switches in classification correctness of a sample
increases with their probability to be mistaken. However, the hardest
samples switch less. (dots individual data, curves means. Skipped
samples above 0.6 are rare and the means are omitted for clarity).
(MNIST, 200 hidden units, trained to 97\% accuracy).}\label{fig:freqhist}
\end{figure}

Next, we tracked the dynamics of the mistakes. Unsurprisingly, mistakes
tend to be most common early in training. The number of newly identified
mistakes drops very rapidly from 90\% to less than 10\%, Fig.~\ref{fig:freqhist}a.
With the start of the 2nd epoch, the rate slows down as mistakes were
already tagged as such, but still some new mistakes are encountered.
Meanwhile the number of update events shows an opposite behavior:
it jumps with the start of the 2nd epoch, as some of the updates are
on re-encountered, now correct samples.

One can wonder whether the memorized mistakes are again misclassified
when re-encountered. \citet{toneva2018empirical} defined forgetting
as the misclassification of a previously correct sample. So-called
unforgettable patterns are learned once and then never misclassified
again. We trained the MNIST network for exactly 5 epochs, reaching
98\% accuracy. As above, most samples are never used for training,
yet still become correct - piggybacking on the learning of related
samples. There were 7190 memorized samples. There are 31 possible
forgetting sequences (ignoring samples without any mistakes), Fig.~\ref{fig:freqhist}b.
The majority of mistakes become correct, most directly (1st pattern
in the list), some with a delay (e.g. 2nd pattern). About 10\% (third
pattern) remain incorrect on the training set. 18.2\% 'forget', i.e.
had a transition from correct to wrong (marked with star), 0.3\% had
2 forgetting events. 

Importantly, the Toneva paper only considers a sample's classification
when it is selected for updating, very sparsely sub-sampling of the
actual sample's trajectory, as the authors acknowledge. Next, we exhaustively
track the error of \emph{all} samples at \emph{every} step. When tracked
this way, samples can switch hundreds of times between correct and
incorrect, Fig.~\ref{fig:freqhist}c top. Thus 'forgetting' is very
common. For mistake samples, the mean number of switches is about
8 times larger (216 vs 34). Switching occurs mostly early in training,
but mistakes keep switching for longer, Fig.~\ref{fig:freqhist}c
bottom. 

The probability to be mistaken across runs (as extracted from multiple
simulations, Fig.~\ref{fig:identity}b) correlates with the number
of switches a sample undergoes, Fig.~\ref{fig:freqhist}d. We interpret
the number of switches as a measure for closeness to the class decision
boundary. Interestingly, the very hardest samples (right-most data
points) experience fewer switches; these samples are supposedly mostly
on the wrong side of the decision boundary, and might not be learned
during training. 

\subsection*{Mistake-gating reduces memory capacity requirements for offline learning}

\begin{figure}
\centering \includegraphics[width=0.6\linewidth]{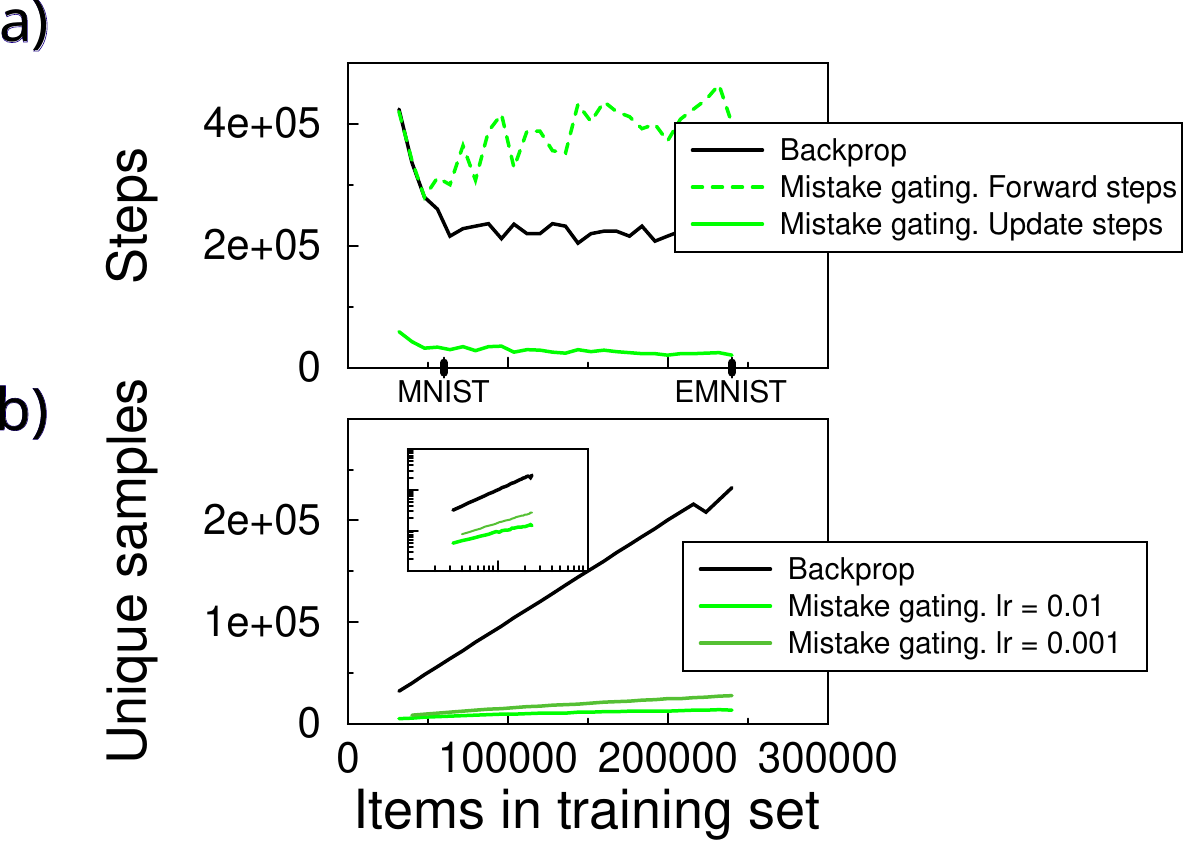}
\caption{Effect of dataset size on mistake gating. The data set size was varied
by using a subset of the Extended MNIST dataset. Networks were trained
until they reached 98\% accuracy on the EMNIST test set.\protect \\
a) The number of forward steps and number of updates. Mistake gated
learning requires more forward steps than backprop (dashed green),
but skips over most so that fewer update steps are needed (solid green).
For backprop (black) the number of update and forward steps is identical.\protect \\
b) Mistake gated learning uses far fewer unique samples than regular
backprop.Inset: same data on a log-log scale. \protect \\
 (200 hidden units, trained to 98\% accuracy).}\label{fig:emnist}
\end{figure}

Next, we examined how the benefits of mistake gating carry over to
larger datasets. This is particularly relevant for continual learning,
as most real-world data is drawn from a much larger dataset. This
is most apparent in natural vision, where scene, projection and lighting
changes can yield practically infinite sample variations. Including
such data variation  improves generalization performance, and data
augmentation is therefore common practice to enlarge datasets \citep{LeCun2015,Kaplan2020}.
Yet, it also exacerbates the inefficiency of standard learning methods.

To examine the dependency on dataset size, we used the extended MNIST
(EMNIST) data set, which contains 240,000 training samples, compared
to 60,000 in MNIST \citep{Cohen2017emnist}. We trained networks on
random subsets of EMNIST and tested on the full EMNIST test set (40000
samples). In this case we trained until it reached an accuracy of
98\%, so that at least one full epoch was required.

Using standard backprop, the number of updates required to reach criterion
varies only weakly with dataset size, Fig~\ref{fig:emnist}a. Indeed,
this is what one would expect if the training subsets sufficiently
cover the full data-set. Mistake gating consistently reduces the number
of updates  to 12\%, irrespective of dataset size.

In addition to the number of updates, a related performance metric
of mistake gating learning is the number of \emph{unique }samples
required for training, i.e. the core-set size. In machine learning
it is common to have unlimited access to the dataset and algorithms
cycle over all data until the required performance is reached. In
biological learning, however, one typically does not have this luxury
and data might need to be stored for offline replay. The hippocampus
is widely considered as an example of such a temporary storage buffer,
which stores events during the day, and during rest and sleep replays
them, consolidating the information into cortical networks. Consistent
with mistake-gated learning, evidence suggests that errors gate hippocampal
memory storage \citep{Sinclair2021}. By identifying the critical
samples, mistake gating reduces the memory requirements for this temporary
buffer.

To study this we assume that all information first has to be stored
in a temporary buffer before it can be committed to the long-term
storage. We stress that this is an extremely abstracted model of consolidation,
ignoring changes in representation, orthogonalization and schema extraction
known to occur in the hippocampus. We also assume the worst case scenario
that no intermediate consolidation events occur, otherwise less memory
would be required.

The number of unique samples needed for training grows with data-set
size, Fig.~\ref{fig:emnist}b. By construction, in regular backprop
the number of unique samples is identical to the dataset size, provided
that more than one training epoch is needed. Storing only incorrect
samples requires however only a fraction of the storage. Interestingly,
the number of unique samples that mistake-gated learning selects to
update from grows sub-linearly in dataset size (light green curve).
A fit yields $\sim S^{0.51\pm0.01}$, where $S$ is the dataset size.
When the learning rate was slowed from a default of $0.01$ to $10^{-3}$,
the exponent was $0.67\pm0.01$ (dark green curve). In summary, while
the number of update steps is approximately invariant across dataset
size, the relative savings in memory requirements increase for larger
datasets. Thus mistake gating allows for more efficient learning of
large datasets, whether occurring naturally or augmented artificially.

\subsection*{Learning from imbalanced datasets}

\begin{figure}
\begin{centering}
\includegraphics[width=0.6\linewidth]{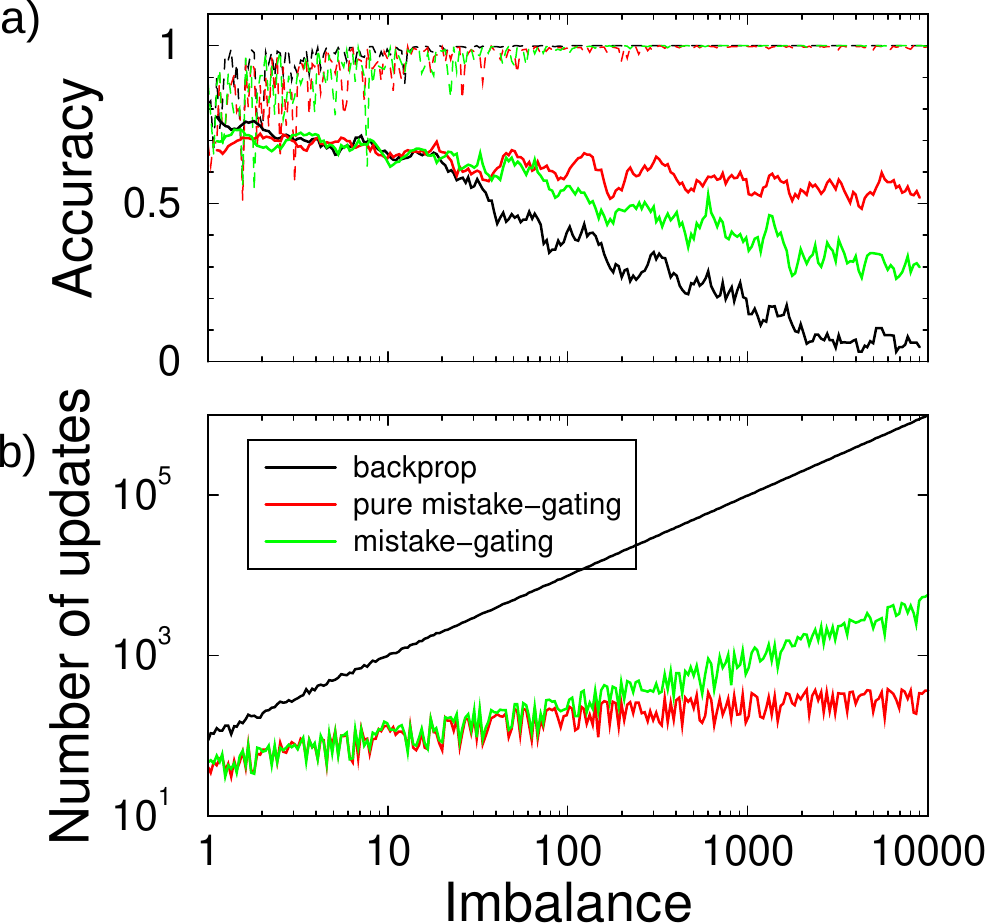}
\par\end{centering}
\caption{Learning from imbalanced data-sets. \protect \\
a) Accuracy of prominent class (dashed top curves) and of minority
classes (bottom curves). Large imbalance impairs performance on minority
classes. Memorized mistake gating partly prevents the decrease (green);
pure mistake gating (red) is most effective. \protect \\
b) The number of applied updates in the prominent class as a function
of imbalance. }\label{fig:imbal}
\end{figure}

In contrast to curated data-sets such as MNIST, many natural datasets
are imbalanced. That is, some classes or data domains are much more
common than others. An example is the over-representation of English
on the Internet. Imbalance is known to impede learning \citep{Johnson2019imbalancereview}.
Typical algorithms counter the imbalance by explicitly weighting the
learning with class probability, however mistake-gating helps to automatically
balance out imbalanced datasets, which is particularly advantageous
for online learning.

We created a version of the MNIST dataset with one prominent digit
class. All but one digit classes had 100 samples. The number of samples
of one prominent digit class was a multiple of the other classes.
 Unsurprisingly, the accuracy on the prominent class quickly reached
perfection, Fig.\ref{fig:imbal}a (dashed curves). However, due to
overwriting, the accuracy on the remaining classes worsened with increasing
imbalance, even though the total number of updates was larger, Fig.\ref{fig:imbal}a
(black curve). With mistake-gating the decrease is substantially less
(green curve) as mistake-gating strongly reduces the number of updates
on the prominent class, Fig.\ref{fig:imbal}b. In this particular
case, pure mistake-gating (red curve) does even better in preventing
the interference. This is partly an artifact due to repeating samples
when the imbalance is above $\sim$60; when these samples are memorized
they trigger updates when re-encountered. In more natural data this
would be much less likely. Alternatively,  an occasional reset of
the memory can be helpful in these scenarios.

\subsection*{Scaling to more complex tasks}

\begin{figure}
\begin{centering}
\includegraphics[width=0.6\linewidth]{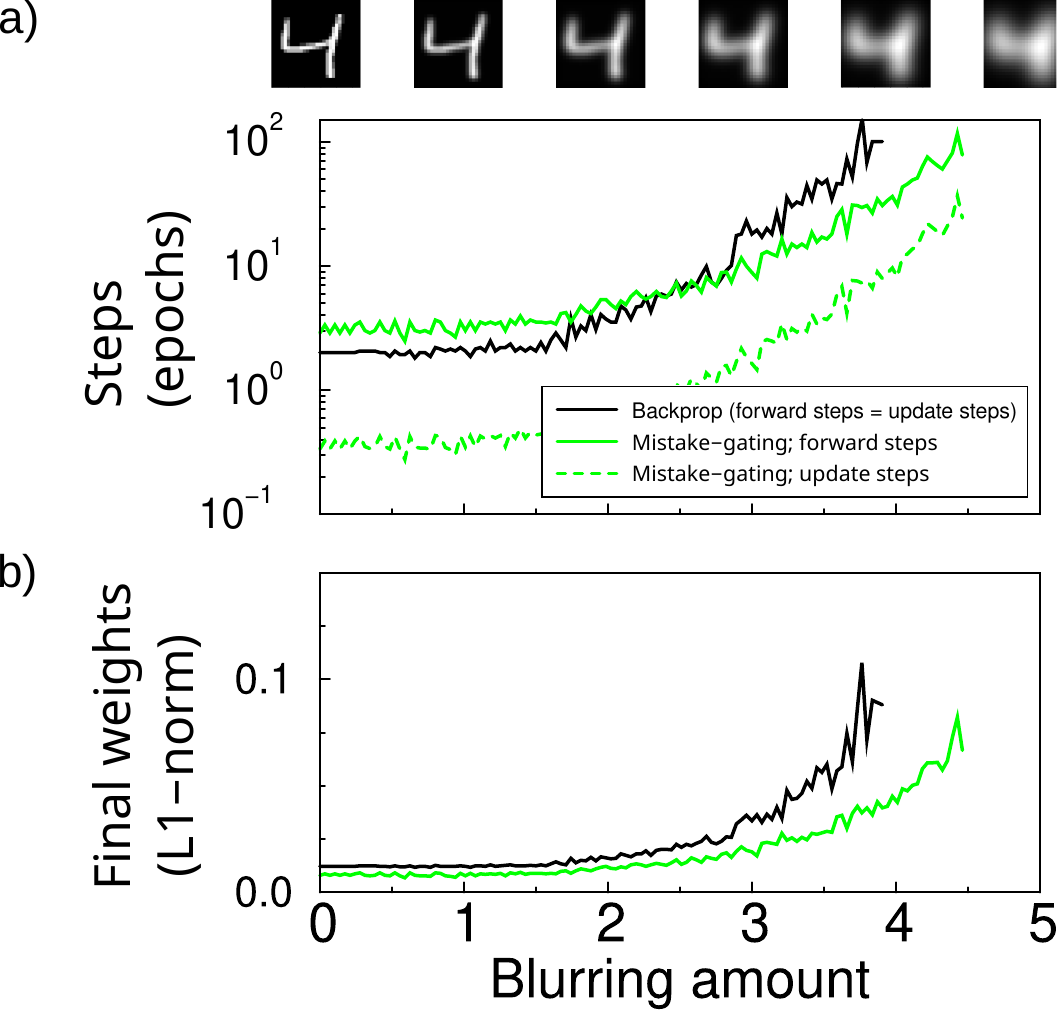}
\par\end{centering}
\caption{Mistake gating for dense, correlated datasets. The standard MNIST
dataset was blurred with a 2D Gaussian filter, the blurring amount
is parameterized as the standard deviation of the filter measured
in pixels.\protect \\
a) The number of passes through the data set (measured in epochs)
to reach 97\% test accuracy increases strongly for blurred images.
Note the log scale. The ratio of update steps of mistake gating learning
compared to backprop is approximately constant.\protect \\
b) The L1 norm of the final weights (measured relative to the initial
condition) pooled over the full network. Mistake-gating leads to smaller
magnitude weights.\protect \\
}\label{fig:blur}
\end{figure}

The MNIST and EMNIST tasks are easy to learn. With a properly tuned
learning rate, high test accuracy on the MNIST task can already be
reached using only single exposure to a part of the training dataset,
i.e.~within one single epoch, Fig.~\ref{fig:trace}. We wondered
whether the savings will diminish when tasks are more difficult and
take many passes over the dataset. To test this, we first parametrically
increased task difficulty by blurring the standard MNIST images with
a 2D Gaussian kernel. After blurring, mean and standard deviation
of the pixel intensities were matched to the original statistics.
Although the blurring is in principle fully invertible, correct classification
requires more precise calculations on pixel intensities and learning
slows down \citep{ahmad2025correlationsruininggradientdescent}. The
number of epochs required increases rapidly for both backprop and
mistake-gated learning as blur increases, Fig.~\ref{fig:blur}a.
However, the number of update steps that memorized mistake gating
requires remains at about 20\% of backprop, Fig.~\ref{fig:blur}a.
This is perhaps surprising: when learning proceeds very slowly, most
samples would be labeled as mistakes. However, further analysis revealed
that already in the first epoch the performance reaches $\sim$90\%,
so mistake-gating still reduces the number of updates. In line with
this, the number of unique samples  tagged as mistakes remains limited,
ranging from $\sim$8000 (out of 60000 samples) without blurring to
$\sim$23,000 at maximum blur (not shown).

Interestingly, while mistake-gating requires more forward steps on
regular MNIST, for the strongest blur the number of forward steps
is actually less. The reason is hard to pin-point exactly, but backprop
updates on all samples some of which become detrimental for learning
under high blur; mistake-gating filters these samples out and becomes
less affected by the noisy descent. Mistake gating finds final weights
with smaller $L_{2}$ and $L_{1}$ norms than backprop, Fig.~\ref{fig:blur}b.
In other words, it has a regularizing effect and we hypothesize that
these smaller norms make learning in later phases easier. 

\begin{figure}
\centering{}\includegraphics[width=14cm]{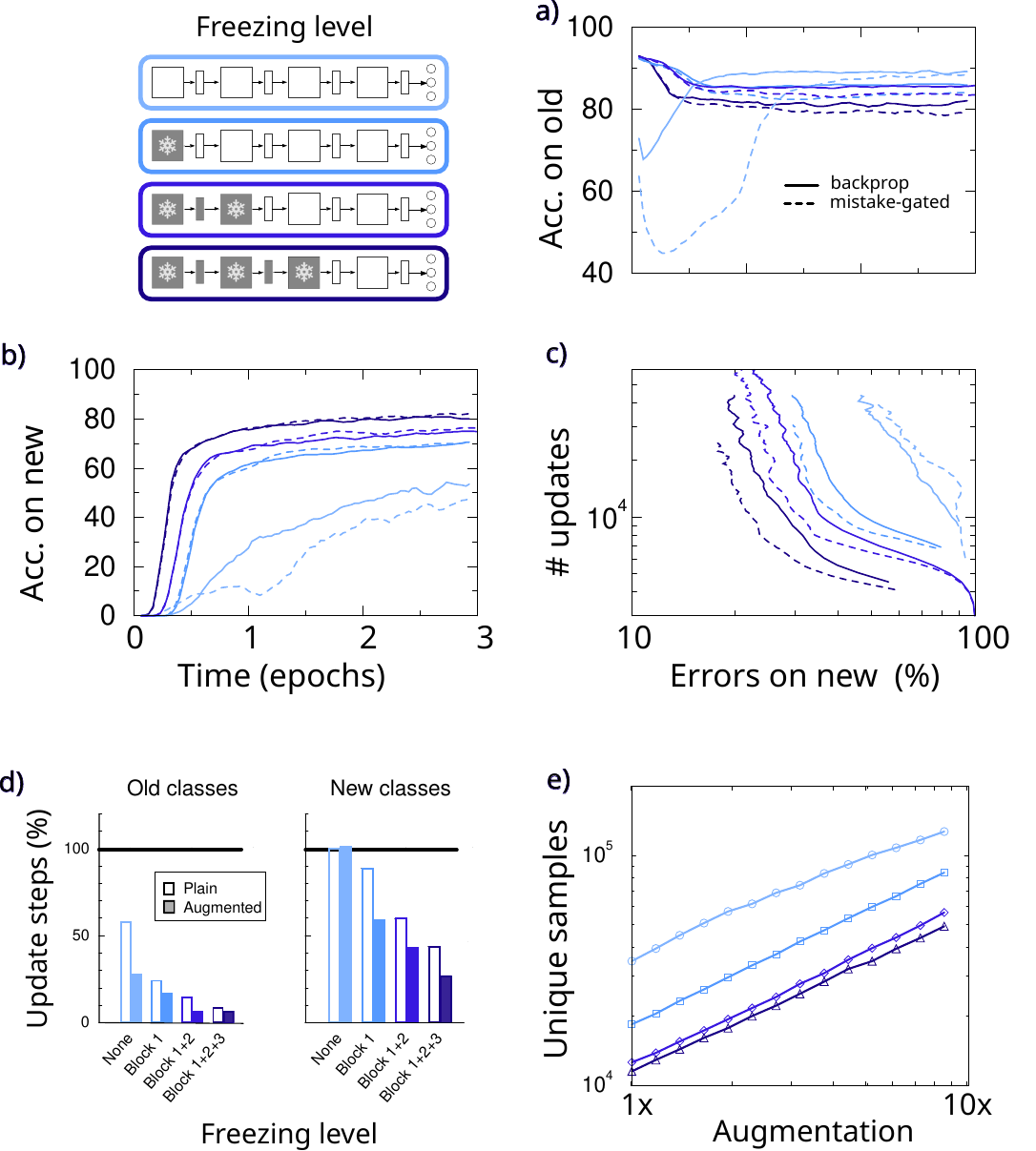}\caption{Mistake gating in incremental learning. Performance on a CIFAR-10
network, pre-trained on 7 classes after which 3 additional classes
are introduced. After the pre-training the network was frozen to
different degrees. Light blue: nothing frozen; darkest blue: all but
last block frozen. \protect \\
a) Test set accuracy on old classes. The most frozen networks maintain
best performance on old classes. Mistake-gating (dashed) used a similar
number of forward steps (epochs) as backprop (solid curves).\protect \\
b) Test set accuracy on new classes, the most frozen networks learned
most quickly to incorporate the new classes. Mistake-gating used a
similar number of forward steps as backprop.\protect \\
c) The number of updates on the new updates required for a certain
test accuracy. Mistake gating (dashed curves) approximately halved
the number of required updates in partly frozen networks, compared
to backprop. \protect \\
d) Plot of saving compared to backprop vs freezing level, showing
most savings for largely frozen networks, and for augmented datasets.
The saving was estimated as the ratio in number of updates at the
highest accuracy levels. \protect \\
e) The number of unique samples (old and new) required to reach 85\%
accuracy scaled sub-linearly with dataset size when using mistake
gating. }\label{fig:cifar}
\end{figure}

\subsection*{Incremental learning on CIFAR}

Next we attempted the more challenging CIFAR-10 dataset, which contains
32$\times$32 RGB images (60,000 train and 10,000 test) of 10 categories
of animals and objects \citep{krizhevsky2009learning}. To mimic hierarchical
visual processing, we used the convolutional Densenet121 network,
which has 121 convolutional layers \citep{huang2018denselyconnectedconvolutionalnetworks}.

When learning from scratch, mistake gating did not yield any savings
(not shown). The reason was that almost all samples were incorrect
in the first epochs, and hence are labeled as mistakes. However, biological
organisms rarely learn complex tasks from a blank slate; instead,
they build upon established sensory representations. We wondered if
the representations that develop in the early layers could be reused
for new categories. We therefore examined an incremental learning
scenario. We first pre-trained the network on 7 out of 10 classes
over 200 epochs (using ADAM, batching, and augmentation) to a test
(training) accuracy of 96.2\% (99.2\%) and recorded the resulting
weights. Subsequently, the remaining 3 categories were included in
the dataset and training was continued on the full dataset (learning
rate 0.001, single batch).

It is known that incremental learning can be faster by only allowing
modification of later layers \citep{yosinski2014transferable,sorrenti2023selective}.
Therefore we restricted the plasticity in varying amounts. Resnet-121
has 4 blocks of layers, and a final linear layer. Plasticity was either
allowed in all layers; allowed in block 2 and beyond; block 3 and
beyond; or, only block 4 and the output layer, Fig.~\ref{fig:cifar}
diagram. Batchnorm layers were always frozen.

The learning curves are strongly dependent on the plasticity restrictions,
Fig.~\ref{fig:cifar}a+b. Interestingly, when just the final block
was plastic (darkest blue), learning was fast and accurate. Instead,
when the network was fully plastic (lightest blue), the introduction
of the new classes knocked down accuracy of the old classes and also
the new classes were learned much more slowly.

What is the effect of mistake-gating in this case? Unlike above, where
mistake-gating typically required more forward steps, here mistake-gating
had virtually no effect on the learning curves of old or new categories;
the dashed and solid curves overlapped, Fig.\ref{fig:cifar}a+b. However,
mistake-gating uses again far fewer updates to reach the same accuracy
level. Unsurprisingly, this is most striking for the old categories
on which the network was already trained, Fig.\ref{fig:cifar}d. However,
also for the new categories it reached similar accuracy using fewer
updates, Fig.\ref{fig:cifar}c+d. We repeated this using ten-fold
augmented data (mirroring and cropping). Training on augmented data
led to relatively larger savings, using a smaller fraction of updates,
Fig.\ref{fig:cifar}d (hatched bars).

Finally, we examined the number of unique samples required to train
to a given test accuracy (85\%) on the new classes. We changed the
dataset size by sampling from the fully augmented data-set. As with
the EMNIST data, we found a sub-linear power-law relation between
dataset size and number of samples used, with respective exponents
0.571 (all plastic), 0.710, 0.717, 0.681 (block 4), Fig.~\ref{fig:cifar}e.
For instance, with just the last block plastic, the number of unique
memorized samples is $56\times10^{3}$, vs $500\times10^{3}$.

In summary, in challenging incremental learning scenarios, mistake-gating
can substantially reduce the number of updates required to learn new
categories provided that the networks are pre-trained. Independently,
incremental learning is much quicker when plasticity is restricted
to later layers (further reducing metabolic cost).

\section*{Discussion}

In summary, when training neural networks, one typically modifies
the parameters for every sample. However, in biological learning plasticity
is sparse, possibly because of the high associated metabolic costs
to it. Mistake-gating can be seen as a temporal sparsification, automatically
selecting the samples for learning in an online manner. Ultimately,
the update inefficiency of standard training methods can be traced
back to the fact that traditional backprop relies on a differentiable,
and hence surrogate, loss function. As a result synapses are updated
regardless of whether the classification is correct or incorrect.
While using the surrogate loss has the benefit of steering the parameters
away from the decision boundaries, and thereby improving generalization
performance, it also means that every sample leads to a parameter
update. This is wasteful. Eventually, late in training most networks
end up in a regime where most samples are correct and mistake-gating
will be highly beneficial. While earlier studies have shown that learning
can indeed be limited to a subset of critical samples, most algorithms
require a fixed data-set and sometimes even a double pass over the
dataset, which is incompatible for the scenarios we considered.

While we concentrated on typical backpropagation learning, mistake-gating
can be used with other, more biological learning algorithms \citep[e.g.][]{sacramento2018dendritic}.
It also naturally combines with curriculum learning \citep{Bengio2009curriculum,9392296},
as well as class imbalanced datasets \citep{Johnson2019imbalancereview},
since rare classes would be classified incorrectly more often.

In terms of experimental predictions, mistake-gating predicts that
memoranda that were once incorrect but now correct, lead to additional
updating. This could be tested by measuring the EEG response to such
stimuli in memory experiments. We predict an error or error-like signal
when such memoranda are encountered. The hippocampus has long been
suggested as a temporary buffer to allow for offline learning. Mistake
gated learning suggests that the content of the hippocampus is carefully
titrated. On one hand, only events that evoke an error signal need
to be stored. On the other hand, such events should be kept in memory,
even if they do not evoke an error anymore, but did so in the past.
Exactly how long the hippocampus should hold on to these patterns,
is a subject for future study and will likely also depend on concurrent
demands.

Related to this, it is interesting to note that traditionally pattern
sparsification is known to increase hippocampal storage \emph{capacity}
\citep{tsodyks88}. Our results with blurred MNIST show that sparse
patterns need less storage for offline learning. Thus pattern sparsification
can also reduce storage \emph{requirements.}

This study is oriented to biological and neuromorphic systems, but
one can wonder whether mistake gating benefits current machine learning.
Using single sample updates, mistake-gating does save CPU time (see
Appendix). In machine learning, however, it is common to use batching.
That is, rapid GPU-based parallel computation of the gradient for
many samples is followed by a single, summed update. Batching thus
also reduces the number of updates. In its basic form mistake gating
does not work well in batched learning, because the probability that
\emph{all} its constituent samples are correct, decreases exponentially
with batch size, nullifying the benefits of mistake-gating learning.
While outside of the scope of the current paper, software engineers
might be able to dynamically construct batches that only contain the
mistaken samples \citep[cf.][]{sathiyanarayanan2025progressivedatadropoutembarrassingly,loshchilov2015online}.
However, this is a tough challenge. The coreset selection methods
in the studies cited above, seem not to be universally used in state-of-the-art
ML, which instead appears to rely on massive uninterrupted data throughput.

In biological systems, a batching-like mechanism can be implemented
by including an additional, transient form of plasticity. Single sample
updates are stored in transient plasticity and consolidated into late-phase
plasticity only once a threshold is reached, an algorithm we termed
synaptic caching \citep{li2020energy}. This saves energy because
transient plasticity is metabolically much cheaper than late phase
plasticity \citep{Mery2005b,Potter2009,Placais2013,amrapali2026mitochondrial}.
Provided that the transient memory decays only when updates are made,
mistake gating reduces the number of consolidation events, compared
to standard backprop with synaptic caching, thus further reducing
energy needs. 

We see mistake-gating as the first line of defense against overzealous
plasticity. Future effort will be to combine the benefits of mistake
learning with competitive updating which reduces the number of updates
in large, over-dimensioned networks \citep{Rossum2024}. Also more
sophisticated sample selection algorithms \citep[e.g.][]{mindermann2022prioritized}
might have biological counterparts. 

\section*{Methods}

Code is available on \url{github.com/vanrossumlab/mistake-gating26}.

\subsection*{Acknowledgments}

It is a pleasure to thank Claudia Danielmeier and Josh Khoo for discussion.
This research was supported by a grant from NVIDIA and some simulations
utilized an NVIDIA RTX A6000.

\subsection*{Appendix }

\subsubsection*{Comparison to \citet{DBLP:journals/corr/abs-1910-00762}}

\begin{figure}
\centering{}\includegraphics[width=8cm]{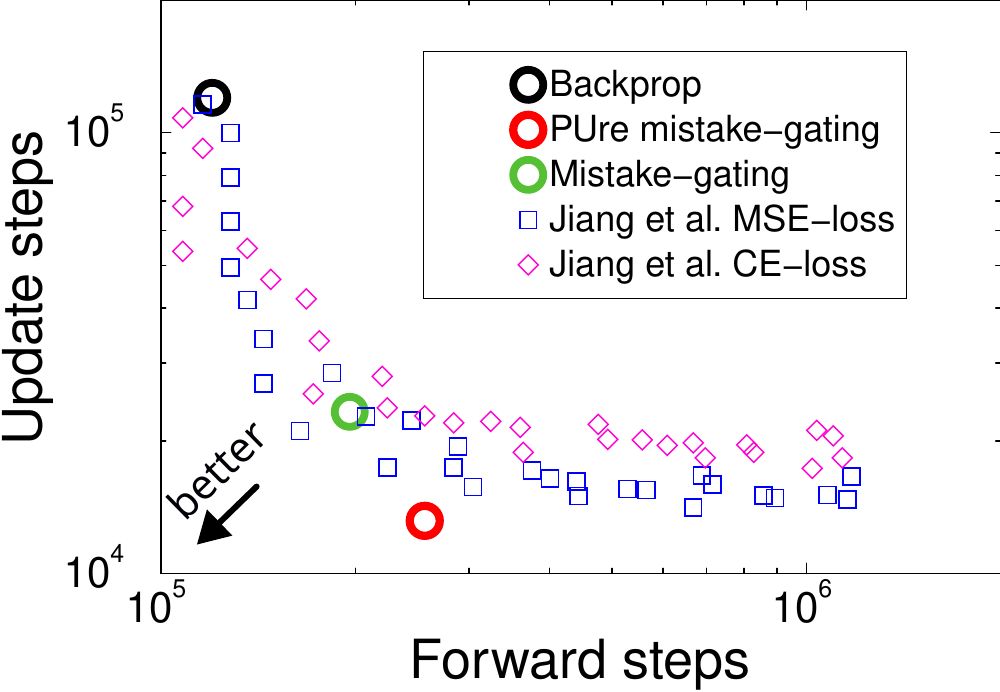}\caption{Mistake-gating performs comparably to \citet{DBLP:journals/corr/abs-1910-00762},
which focusses learning on samples with highest (surrogate) loss.
Their selectivity parameter $\beta$ was varied from 0 (unselective;
top left) to about 5 (highly selective). This trades off number of
update steps vs number of forward steps. Mean squared error loss (blue)
leads to a slightly better trade-off than cross-entropy loss (purple).
Also shown: standard backprop (black), pure mistake-gating (red),
and mistake-gating (green).  (MNIST, 200 hidden units, trained to
97\% test accuracy; the time-window of the loss history used to calculate
relative loss was unlimited)}\label{fig:jiang}
\end{figure}

A number of papers have proposed to focus learning towards critical
samples. Selection is typically based on the actual surrogate loss,
rather than correct classification used here. We compared mistake-gating
to the method of sample selection proposed by \citet{DBLP:journals/corr/abs-1910-00762}.
The losses of recent samples (stored in a history buffer) are ranked
into a cumulative distribution $P$, and the probability for an update
is set to $P^{\beta}$, where $\beta$ is a free parameter. When $\beta=0$,
all samples will be updated. For larger $\beta$, the samples with
higher losses are more likely to lead to an update; the sample with
highest loss with always trigger an update. The largest reduction
in the number of updates occurs when most selective (i.e. large $\beta$)
and longest history buffer, however, as in our work, it slows down
the learning requiring more forward passes. Our method performs comparable
to this method, Fig.~\ref{fig:jiang}. Arguably, mistake-gating is
more straightforward to implement biologically, requiring no memory
of the history of relative losses.

\subsubsection*{Benchmarking}

\begin{figure}
\centering{}\includegraphics[width=8cm]{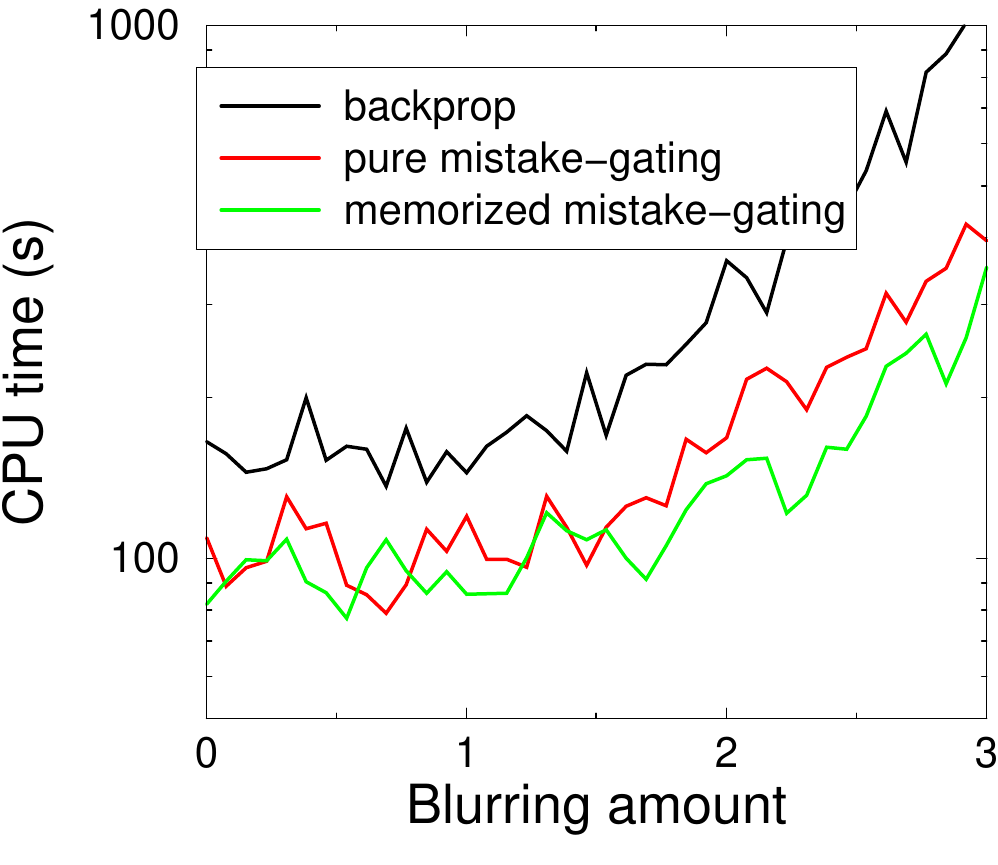}\caption{CPU time for training blurred MNIST for the various algorithms. The
value at zero corresponds to standard MNIST. The fluctuations are
mostly due to the noisy nature of reaching criterion performance.
 (200 units, 97\% test accuracy criterion).}\label{fig:bench}
\end{figure}

Our algorithm is not particularly efficient on GPUs as it has not
been designed for batching, but on single batch CPU training, it can
save compute time. We compared the wall compute time of the various
algorithms on MNIST and blurred MNIST, excluding the substantial time
taken to measure test-set performance. Tests were done on a workstation
with AMD 3970X processor with the training on a single CPU thread.
We used the blurred MNIST protocol, Fig.~\ref{fig:bench}. We found
that feedforward steps took $(18.7\pm0.1)$s per epoch, while an epoch
worth of backprop steps took $(51.7\pm0.1)$s, generally consistent
with the finding that backprop steps require $2\ldots3\times$ more
compute than forward steps. As a result, although mistake gating has
more forward steps, it uses only half the CPU time for standard MNIST.
As the blurring increases the reduction in CPU time increases to about
4 times. Interestingly, pure mistake gating takes a bit longer than
memorized mistake gating as it requires more feed-forward steps, but
only slightly fewer updates.

\end{document}